\documentclass[conference,10pt]{IEEEtran}
\usepackage{algpseudocode}                                 
\usepackage{algorithm}
\usepackage{graphicx}                                      
\usepackage{amsmath}
\usepackage{amssymb}
\usepackage{amsthm}
\usepackage{amsfonts}
\usepackage{booktabs}
\usepackage{multirow}
\usepackage[subrefformat=parens,farskip=0pt,justification=centering]{subfig}
\usepackage{color}
\usepackage{cite}                                          
\usepackage{comment}                                       
\usepackage{soul}                                          
\soulregister\cite7
\soulregister\ref7
\soulregister\pageref7
\usepackage{amsthm}
\usepackage{etoolbox}                                      
\usepackage{url}
\usepackage{nth}                                           
\usepackage{bm}                                            
\usepackage{courier}
\usepackage{balance}
\usepackage{threeparttable}
\usepackage[bookmarks=false]{hyperref}
\hypersetup{
    colorlinks = true,
    citecolor  = blue,
    linkcolor  = blue,
    urlcolor   = blue,
}
\usepackage{tikz}
\usetikzlibrary{positioning, fit}
\usepackage{filecontents}                                  
\usepackage{pgfplots}
\usepackage{pgfplotstable}
\pgfplotsset{compat=newest}
\usepackage{caption}
\usepackage{cleveref}
\Crefformat{figure}{Fig.~#2#1#3}                           
\Crefname{subfigure}{Fig.}{Figs.}
\Crefformat{table}{TABLE~#2#1#3}                           
\definecolor{CUHKorange}{RGB}{244,106,18} 
\definecolor{CUHKblue}{RGB}{0,111,190}    
\definecolor{CUHKgreen}{RGB}{0,127,128}   
\definecolor{CUHKred}{RGB}{228,46,36}     
\definecolor{CUHKyellow}{RGB}{198,148,34} 
\definecolor{CUHKdark}{RGB}{114,44,114}   
\definecolor{CUHKmiddle}{RGB}{144,44,144} 
\definecolor{CUHKlight}{RGB}{167,44,167}

\newtheorem{mytheorem}{\textbf{Theorem}}

\renewcommand{\vec}[1]{\boldsymbol{#1}}    
\algrenewcommand\textproc{\texttt}

\makeatletter
\let\OldStatex\Statex
\renewcommand{\Statex}[1][3]{%
  \setlength\@tempdima{\algorithmicindent}%
  \OldStatex\hskip\dimexpr#1\@tempdima\relax
}
\makeatother

\RequirePackage[normalem]{ulem} 
\RequirePackage{color}\definecolor{RED}{rgb}{1,0,0}\definecolor{BLUE}{rgb}{0,0,1} 

\graphicspath{{./figs/}}

\IEEEoverridecommandlockouts

\newcommand{\norm}[1]{\left\lVert#1\right\rVert}

\definecolor{myorange}{RGB}{238,97,42}  %
\definecolor{myblue}{RGB}{178,179,249}  
\definecolor{mygrey}{RGB}{166,166,166}  %
\definecolor{mygreen}{RGB}{180,210,36}  
\definecolor{myred}{RGB}{238,0,0}       
\definecolor{myyellow}{RGB}{198,148,34} 
\definecolor{mydark}{RGB}{114,44,114}   
\definecolor{mymiddle}{RGB}{144,44,144} 
\definecolor{mylight}{RGB}{167,44,167}  

\begin{document}

\title{
	A Unified Approximation Framework for Compressing and Accelerating Deep Neural Networks
	\thanks{
        Corresponding authors: Wenjian Yu (\texttt{yu-wj@tsinghua.edu.cn}) and Bei Yu (\texttt{byu@cse.cuhk.edu.hk}).
	}
}

\author{
    \IEEEauthorblockN{
        Yuzhe Ma\IEEEauthorrefmark{1},
        Ran Chen\IEEEauthorrefmark{1},
        Wei Li\IEEEauthorrefmark{1},
        Fanhua Shang\IEEEauthorrefmark{2},
        Wenjian Yu\IEEEauthorrefmark{3},
        Minsik Cho\IEEEauthorrefmark{4},
        Bei Yu\IEEEauthorrefmark{1},
    }
    \IEEEauthorblockA{
        \IEEEauthorrefmark{1}CSE Department, Chinese University of Hong Kong, \\
        \IEEEauthorrefmark{2}School of Artificial Intelligence, Xidian University, \\
        \IEEEauthorrefmark{3}BNRist, Dept.~Computer Science \& Tech., Tsinghua University, \ \ 
        \IEEEauthorrefmark{4}IBM T.~J.~Watson
    }
}
\maketitle
\thispagestyle{empty}

\begin{abstract}

Deep neural networks (DNNs) have achieved significant success in a variety of real world applications, i.e., image classification. 
However, tons of parameters in the networks restrict the efficiency of neural networks due to the large model size and the intensive computation. 
To address this issue, various approximation techniques have been investigated, which seek for a light weighted network with little performance degradation in exchange of smaller model size or faster inference.
Both low-rankness and sparsity are appealing properties for the network approximation.
In this paper we propose a unified framework to compress the convolutional neural networks (CNNs) by combining these two properties, while taking the nonlinear activation into consideration.
Each layer in the network is approximated by the sum of a structured sparse component and a low-rank component, which is formulated as an optimization problem. Then, an extended version of alternating direction method of multipliers (ADMM) with guaranteed convergence is presented to solve the relaxed optimization problem. 
Experiments are carried out on \emph{VGG-16}, \emph{AlexNet} and \emph{GoogLeNet} with large image classification datasets. 
The results outperform previous work in terms of accuracy degradation, compression rate and speedup ratio.
The proposed method is able to remarkably compress the model (with up to $4.9\times$ reduction of parameters) at a cost of little loss or without loss on accuracy.


\end{abstract}

\section{Introduction}

As neural networks become deeper and deeper, the representation ability of neural network keeps improving, leading to significant performance promotion in a variety of tasks.
However, the model size and the computation cost of neural networks are also increasing due to the huge amount of weights learned, which results in low throughput in inference stage and restrains the deployment  on resource-limited systems.
For example, embedded devices may lack enough storage and computation power to execute the giant networks.
Meanwhile, deep neural networks are demonstrated to be over-parameterized \cite{SPEED-NIPS2013-Denil}, which motivates researchers to explore efficient approaches to make the deep models compact.

\begin{figure}[!tb]
	\centering
	\subfloat[]{\includegraphics[width=1.0\linewidth]{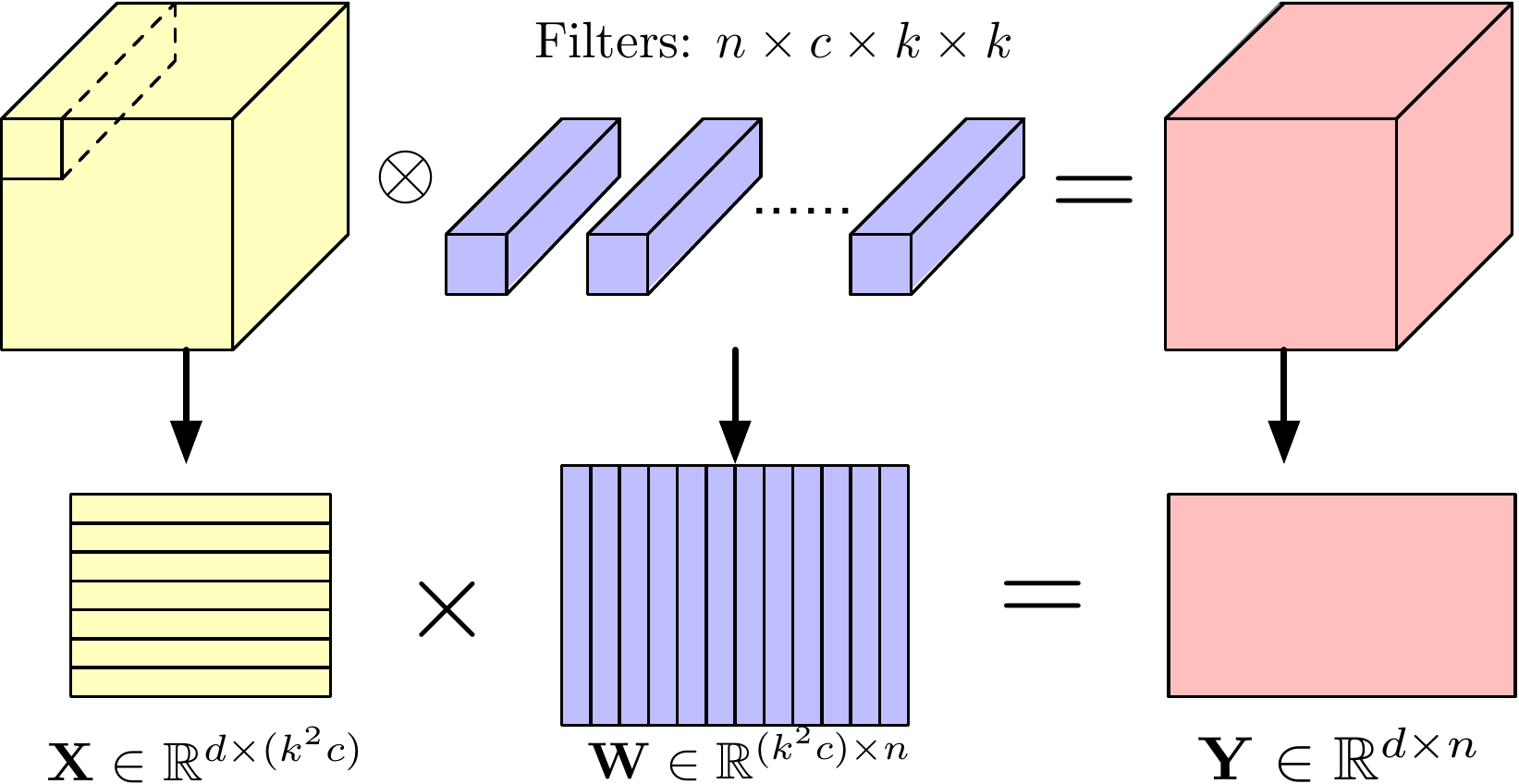} \label{fig:network-im2col}} \\
	\subfloat[]{\includegraphics[height=2.0cm]{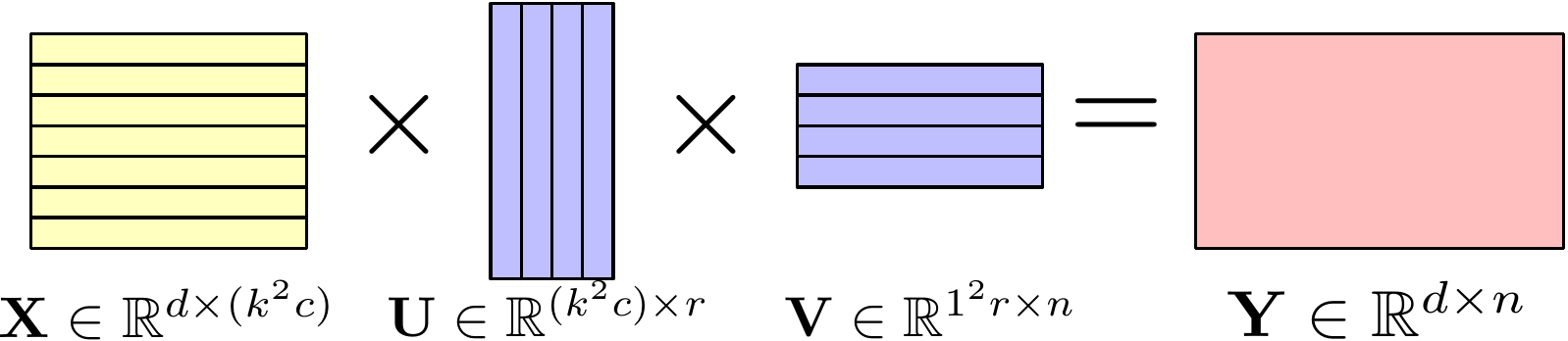} \label{fig:network-low-rank}}      \\
	\subfloat[]{\includegraphics[height=2.0cm]{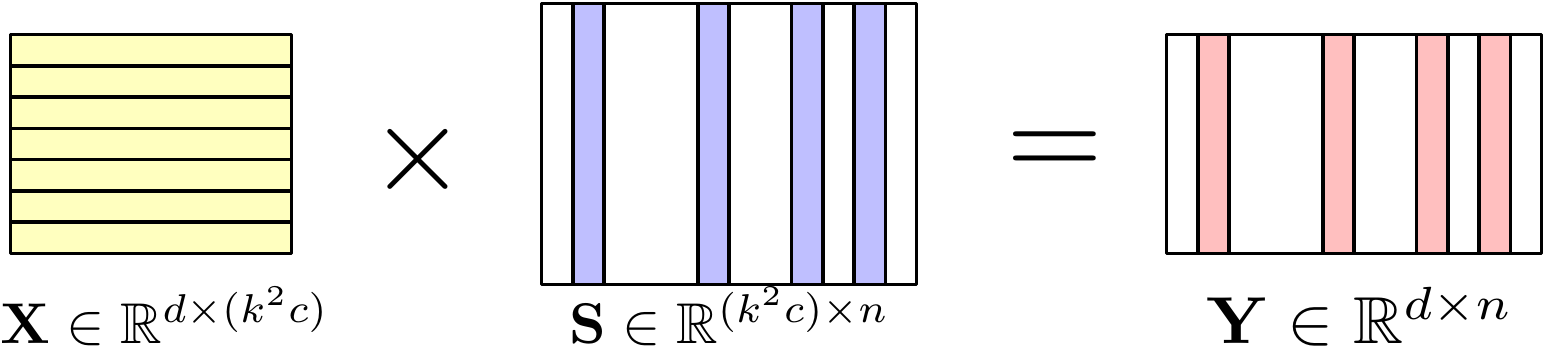} \label{fig:network-sparse}}
	\caption{(a) Transform convolution to matrix multiplication;
		(b) Approximate weight matrix $\mathbf{W}$ using two matrices with lower-rank;
		(c) Impose structured sparsity on weight matrix $\mathbf{W}$.}
	\label{fig:network}
\end{figure}

Approximating the deep models involves removing the redundancy and seeking for simplified structures such that the network after approximation may retain the performance on original tasks. 
Low-rankness and sparse connection are the most commonly applied assumptions when approximating a model.
The illustration is presented in \Cref{fig:network}.
Sparse connection can be realized by pruning a pre-trained network, which is the most straightforward approach.
A hard thresholding approach is proposed in \cite{SPEED-ICLR2016-Han}, which achieves high sparsity by removing the weights with less importance.
Structured sparse connections can be learned by imposing group sparse regularizations during the training \cite{SPEED-NIPS2016-Wen},
as shown in \Cref{fig:network-sparse}. 
It also favors computation acceleration. 
In additional to sparse connection, low-rankness is another desired structure for model approximation.
The weight matrices in the neural networks of low-rank can be further decomposed into smaller matrices, so as to reduce the amount of parameters as well as the computation cost \cite{SPEED-IJCNN2018-Dai,SPEED-CVPR2015-Zhang}, as shown in \Cref{fig:network-low-rank}. 
Tensor decomposition is applied in \cite{SPEED-NIPS2015-Novikov}, where the weight tensor in fully-connected (FC) layer is approximated by a series of smaller kernels. 
Similar to sparse connection networks, the low-rank filter of neural network can also be learned by imposing regularizations \cite{SPEED-ICLR2016-Tai,SPEED-NIPS2017-Alvarez}.
An intuitive extension of these work is to consider both low-rank structure and sparse structure simultaneously. 
In \cite{SPEED-CVPR2017-Yu}, a layer in the pre-trained neural network is decomposed into a low-rank component and a sparse component by a greedy algorithm.
The obtained networks are compressed but not accelerated due to the non-structured sparse weights.

Performing approximation or pruning to a pre-trained network may inevitably result in performance loss.
In order to retain the accuracy of a pre-trained model, some approaches aim to minimize the reconstruction error of the feature maps in each layer through solving an optimization problem with specified constraints on the rank or sparsity of the filters.
The reconstruction error is measured between the linear response in original network and approximated one \cite{SPEED-CVPR2017-Yu,SPEED-ICCV2017-He}.
Since the non-linearity such as  Rectified Linear Units (ReLU) \cite{DL-ICML2010-Nair} follows the linear filters in most neural networks, only the error of positive response is accumulated and  the error of negative response is omitted, which makes the accuracy more dependent on the positive response reconstruction.
In \cite{SPEED-CVPR2015-Zhang}, a method for reconstructing non-linear response is proposed.

In this work, we propose a unified approximation framework for CNNs which approximates the convolutional layers with two components, including a structured sparse component and a low-rank component. 
In contrast to \cite{SPEED-CVPR2017-Yu}, our constraints not only facilitate model compression but also favors acceleration because of the structured sparse weights \cite{SPEED-NIPS2016-Wen}.
We retain the accuracy of the model by approximating the nonlinear response after activation. 
The layer-wise network approximation problem is formulated as minimizing the reconstruction error of the response after non-linear ReLU. 
To overcome the resulted difficulty of non-convex optimization, we propose a convex relaxation scheme which considers the constraints for structured sparsity and low-rankness, and then solve it with an extension of alternating direction method of multipliers (ADMM) \cite{OPT-FTML2011-Boyd}. 
Moreover, we prove that the extended ADMM algorithm converges to the optimal solution of the relaxed problem.



The proposed method is evaluated on well-known DNN architectures, including \emph{VGG-16} \cite{IMGC-ICLR2015-VGG}, \emph{NIN} \cite{IMGC-ICLR2014-NIN}, \emph{AlexNet} \cite{IMGC-NIPS2012-AlexNet} and \emph{GoogLeNet} \cite{IMGC-CVPR2015-GoogleNet}.
For \emph{VGG-16} with the CIFAR-10 dataset, we achieve $4.4\times$ model compression with only 0.4\% accuracy drop. Meanwhile, with the compressed model the inference is accelerated by $2.2\times$. 
For \emph{AlexNet} with the ImageNet dataset, we achieve $4.9\times$ model compression at the cost that the top-5 accuracy drops slightly from 81.3\% to 80\%. 
For \emph{GoogLeNet} with the ImageNet dataset, the proposed method also brings $2.9\times$ reduction of the model parameters without any degradation on the accuracy of inference. 
These experimental results reveal that the proposed approximation framework is able to remarkably compress the CNN models while keeping high accuracy.

The rest of this paper is organized as follows. 
In \Cref{sec:related}, related literature on DNN compression and acceleration is summarized. 
The problem formulation of the proposed methodology is given in \Cref{sec:problem}. 
\Cref{sec:alg} presents a numerical optimization algorithm for solving the problem. 
The experimental results are reported in \Cref{sec:results}, and \Cref{sec:conclu} concludes the paper.

\section{Related Work}
\label{sec:related}

\textbf{Neural Network Sparsification}.
Despite the appealing performance of the deep neural network, it has been demonstrated that there is much redundancy which leads to computation overhead and large model size.
Therefore, sparsifying some over-parameterized layers in neural network is a straightforward method to eliminate the redundancy while preserving the performance.
The majority of the parameters in a sparse layer are zeros, thus the parameters can be stored with compressed representation, e.g., compressed sparse row (CSR) format, for size reduction.
A three-stage pipeline is proposed in \cite{SPEED-ICLR2016-Han}. 
The parameters that are smaller than a threshold are considered as less important and are set to zeros.
Then retraining is performed on the sparse structure to restore the accuracy. 
However, the sparse pattern is non-structured which has limited benefit for speedup during inference due to the poor weight locality.
\cite{SPEED-ICLR2017-Li} proposes to prune the entire convolution kernel rather than single element based on the intensity. 
A structured sparse learning algorithm is proposed in \cite{SPEED-NIPS2016-Wen}, which enables to learn a network with structured sparse network by applying group sparse regularizations during training.
Since structured sparsity leads to zero-columns and zero-rows in the lowered matrices, \cite{SPEED-NIPS2016-Wen} further proposes to reduce the dimension of lowered matrices by removing these zero-columns and zero-rows, which reduces the dimension of the lowered weight matrix when applying General Matrix-Matrix Multiplication (GEMM) function and accelerates inference. 
A channel pruning method is proposed in \cite{SPEED-ICCV2017-He}, which can be considered as a special case of structured sparsity.
The difference is that channel pruning is performed on a pre-trained model rather than training the model from scratch.
\cite{SPEED-ICCV2017-He} formulated the problem as $l_0$-norm minimization problem, trying to find the ``informative'' channels of the feature map and the corresponding weights. 
Instead of trying to minimize the reconstruction error layer by layer, \cite{SPEED-CVPR2018-Yu} targets at a unified goal which is to minimize the reconstruction error of important response in the final response layer. 
In this paper, we are more interested in exploring structured sparsity since it not only facilitates compressing the networks but also acceleration.

\textbf{Low-Rank Approximation}.
In addition to sparsifying a network, low-rank approximation is another sort of approach which can be applied for both network compression and acceleration.
In modern convolutional neural networks (CNNs) structure, filters are usually a 4-D tensor.
Some tensor decomposition techniques are leveraged for acceleration and compression. 
A straightforward idea is to replace the 4-D tensor with two consecutive tensors with lower-rank \cite{SPEED-BMVC2014-Jaderberg}.
In addition, other kinds of tensor decomposition can also be applied. 
In \cite{SPEED-NIPS2015-Novikov}, fully-connected layers are converted to the Tensor Train format, resulting in compression by a huge factor.
CP-decomposition of the filter tensors is proposed in \cite{SPEED-ICLR2015-Lebedev}. 
A relevant approach to low-rank approximation is tensor sketching \cite{SPEED-IJCAI2018-Kas}.
The difference is that low-rank approximation will increase the network depth since an original layer will be decomposed into multiple layers. 
In order to conduct low-rank approximation more efficiently, methods for training neural networks with low-rank filters are investigated \cite{SPEED-ICLR2016-Tai,SPEED-NIPS2017-Alvarez,SPEED-IJCNN2018-Dai,SPEED-ICCV2017-Wen}.

\begin{figure*}[!tb]
	\centering
	\includegraphics[width=.68\linewidth]{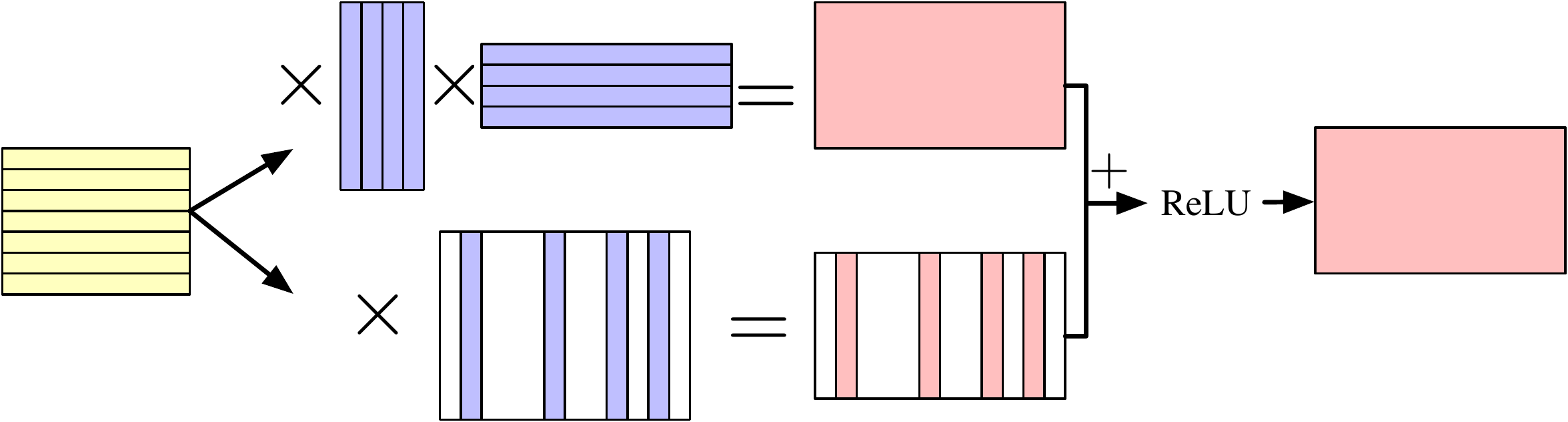}
	\caption{Structure combining sparse and low-rank decomposition.}
	\label{fig:structure}
\end{figure*}

Most of those low-rank approximation-based methods for a pre-trained model like \cite{SPEED-BMVC2014-Jaderberg,SPEED-NIPS2014-Denton} consider reconstructing the response of linear block of a network, while ignoring the following non-linear activation function like ReLU \cite{DL-ICML2010-Nair} which is widely applied in DNNs.
A method for low-rank approximation of non-linear response in convolutional networks is proposed in \cite{SPEED-CVPR2015-Zhang}, which is demonstrated to have more speedup than approximating linear-response only.

Although both low-rank approximation and network sparsification are appealing, there are not so much work that consider how to combine them.
Some previous work aims to train a network with both group-sparse and low-rank regularizations \cite{SPEED-NIPS2017-Alvarez,SPEED-ECCV2016-Zhou}, which impose the constraints of low-rank and sparse on the filters at the same time.
Considering that filters tend to be both low-rank and sparse, a layer in a pre-trained DNN is approximated by the sum of a non-structured sparse component and a low-rank component for compression in \cite{SPEED-CVPR2017-Yu}.
Similar to \cite{SPEED-BMVC2014-Jaderberg}, it relies on reconstructing the linear response of a layer to constrain accuracy loss. 
A potential problem is that non-structured sparsity may not favor computation acceleration well.

Although there are existing work combining the low-rank approximation and network sparsification and the work approximating the non-linear response of network, no one has combined all the ideas together. 
In this work, we propose to approximate the layers in CNN with the sum of a structured sparse component and a low-rank component, and minimize the reconstruction error, while taking the non-linear activation into account. 
It results in a unified approximation framework for compressing and accelerating DNN models.

\section{Problem Formulation}
\label{sec:problem}
In this section, we introduce our mathematical formulation for network approximation using structured sparse and low-rank decomposition, while taking non-linearity into account.
To this end, we propose to formulate the problem into a unified optimization model.
In the following context, we focus on CNNs which involve a large model size. 

In an FC layer of a CNN, the output feature map can be computed as

\begin{equation}
	\vec{Y} = \vec{W}\vec{X},
	\label{eq:mm}
\end{equation}
where $\vec{X} \in \mathbb{R}^{m}$ and $\vec{Y} \in \mathbb{R}^{n}$ represent the input feature vector and output response, respectively.
$\vec{W} \in \mathbb{R}^{n \times m}$ denotes the weight matrix.
For a convolutional layer the convolution operation can also be represented as \Cref{eq:mm}. 
The illustration is shown in \Cref{fig:network-im2col}. 
The convolution filter is $\mathcal{W} \in \mathbb{R}^{n\times c\times k \times k}$, where $k$ is spatial size, $c$ is the number of input channels and $n$ is the number of filters.
The filter can be reshaped to a matrix with size $n$-by-$k^2c$.
The input $\vec{X}$ is lowered to a matrix such that each $k^2 c$ volume involved in a convolution forms a column. 
Then the convolution operation is converted to matrix multiplication.

The information loss is inevitable when we approximate the original filters by low-rank or sparse filters, which may cause performance degradation.
In order to compress the network and accelerate the computation, we perform low-rank approximation and network sparsification simultaneously.
The output feature maps of a layer is generated by the sum of convolving with each filter.
In order to preserve the performance, we aim at minimizing the reconstruction error of the response generated by the approximated filters in each layer after activation. 
An example block structure in the network is demonstrated in \Cref{fig:structure}. 
Then, the problem is formulated as follows:

\begin{align}
    \begin{aligned}
    \label{eq:origin}
    \min_{\vec{A},\vec{B}} \ &\sum_{i=1}^{N} \norm{\vec{Y}_i - r((\vec{A} + \vec{B})\vec{X}_i)}_F^2,  \\
    \textrm{s.t.} \ \ 
    & \norm{\vec{A}}_0 \leq S,  \quad \text{rank}(\vec{B}) \leq L.
    \end{aligned}
\end{align}

Here $\vec{Y}_i$ and $\vec{X}_i$ represent the output feature map and the input feature map of a layer, respectively.
Structured sparse component $\vec{A}$ and low-rank component $\vec{B}$ are two weight matrices we are looking for, each of which is the lowered matrices of a 4-D tensor.
$N$ is the total number of samples used for approximation.
$\norm{\cdot}_F$ is Frobenius norm.
$r(\cdot)$ is the activation function in the network, i.e., ReLU($\cdot$). 
$S$ and $L$ are user-defined target sparsity level and target rank for the filters.

\section{Optimization Methodology}
\label{sec:alg}

\subsection{Problem Relaxation}

Solving Problem \eqref{eq:origin} directly involves both $l_0$ minimization and rank minimization, which is NP-hard. 
Besides, we want $\vec{A}$ to be structured sparse, which leads to extra difficulty. 
To tackle this challenge, we apply convex relaxation to the constraints.
The rank constraint on $\vec{B}$ is relaxed by nuclear norm of $\vec{B}$, which is the sum of the singular values of $\vec{B}$.
As for the $l_0$ norm constraints, a general way is to relax it by $l_1$ norm which is convex and has good performance in imposing sparsity.
However, as we discussed above, structured sparse patterns can be more easily used for computation acceleration. 
Therefore, here we relax $l_0$ constraint by $l_{2,1}$ norm (the sum of the Euclidean norms of the columns) such that the zero elements in $\vec{A}$ appear column-wise. 
Then, the original problem is reformulated as 

\begin{equation}
\min_{\vec{A},\vec{B}} \sum_{i=1}^{N} \norm{\vec{Y}_i - r((\vec{A} + \vec{B})\vec{X}_i)}_F^2  + \lambda_1\norm{\vec{A}}_{2,1}+\lambda_2\norm{\vec{B}}_{*}, \label{eq:relaxation}
\end{equation}
where $\norm{\cdot}_{2,1}$ is $l_{2,1}$ norm and $\norm{\cdot}_{*}$ is nuclear norm.
$\lambda_1$ and $\lambda_2$ are coefficients of the relaxed terms. 
The problem \eqref{eq:relaxation} now is a convex optimization problem. 
To solve it, we make use of the alternating direction method of multipliers (ADMM), 
which is widely used in large-scale problems arising in statistics \cite{OPT-FTML2011-Boyd}.
Especially, the optimal solution of the sub-problems involving $l_{2,1}$-norm and nuclear norm  can be obtained in closed-form as in subspace learning \cite{ML-TPAMI2013-Liu} and the singular value thresholding (SVT) operator \cite{OPT-SIOPT2010-Cai}, respectively. 

By introducing an auxiliary variable $\vec{M}$, 
the Problem \eqref{eq:relaxation} can be rewritten as
\begin{align}
    \begin{aligned}
    \min_{\vec{A},\vec{B},\vec{M}} & \sum_{i=1}^{N}\norm{\vec{Y}_i - r(\vec{M}\vec{X}_i)}_F^2  + \lambda_1\norm{\vec{A}}_{2,1}+\lambda_2\norm{\vec{B}}_{*},  \\
    \textrm{s.t.} \ \ 
    & \vec{A} + \vec{B} = \vec{M}. \label{eq:3-vars}
    \end{aligned}
\end{align}
Then the augmented Lagrangian function of Problem \eqref{eq:3-vars} is
\begin{align}
    \begin{aligned}
    & L_t(\vec{A},\vec{B},\vec{M},\vec{\Lambda})
      = \sum_{i=1}^{N}\norm{\vec{Y}_i - r(\vec{M}\vec{X}_i)}_F^2  + \lambda_1\norm{\vec{A}}_{2,1}  \\
    & \ + \lambda_2\norm{\vec{B}}_{*} + \langle \vec{\Lambda}, \vec{A}+\vec{B}-\vec{M} \rangle + \frac{t}{2}\norm{\vec{A}+\vec{B}-\vec{M}}_F^2,
    \end{aligned}
\end{align}
where $t > 0$ is the penalty parameter and $\vec{\Lambda}$ is Lagrange multiplier.
$\langle \cdot, \cdot \rangle$ represents the inner product operator.

\subsection{Variables Update}
ADMM solves the minimization problem of $L_t(\vec{A},\vec{B},\vec{M},\vec{\Lambda})$ iteratively.
The variables are alternatively updated in each iteration. 
To update $\vec{A}, \vec{B}, \vec{M}$ in iteration $k+1$, our algorithm takes two steps.
Firstly,  we consider the following three sub-problems.

\begin{align}
    \underset{\vec{A}}{\textrm{min }} &\lambda_1\norm{\vec{A}}_{2,1} + \frac{t}{2}\norm{\vec{A}+\vec{B}_k-\vec{M}_k + \frac{\vec{\Lambda}_k}{t}}^2_F, \label{eq:sub-A}\\
    \underset{\vec{B}}{\textrm{min }} &\lambda_2\norm{\vec{B}}_{*} + \frac{t}{2}\norm{\vec{B}+\hat{\vec{A}_{k}} -\vec{M}_k + \frac{\vec{\Lambda}_k}{t}}^2_F, \label{eq:sub-B}\\
    \underset{\vec{M}}{\textrm{min }} &\sum_{i=1}^{N}\norm{\vec{Y}_i - r(\vec{M}\vec{X}_i)}_F^2 + \langle \vec{\Lambda}_k, \hat{\vec{A}_{k}} +\hat{\vec{B}_{k}} -\vec{M} \rangle \nonumber \\
     &+ \frac{t}{2}\norm{\hat{\vec{A}_{k}} +\hat{\vec{B}_{k}} -\vec{M}}_F^2. \label{eq:sub-M}
\end{align}

All these three problems are proximal mapping problems.
For Problem \eqref{eq:sub-A}, the optimal solution is given by

\begin{equation}
    \hat{\vec{A}_{k}}  = \textrm{prox}_{\frac{\lambda_1}{t}\norm{\cdot}_{2,1}}(\vec{M}_k -\vec{B}_k - \frac{\vec{\Lambda}_k}{t}). \label{eq:A-prox}
\end{equation}

The explicit representation of \Cref{eq:A-prox} can be derived based on \cite{ML-TPAMI2013-Liu}.
Let $\vec{C} = \vec{M}_k -\vec{B}_k - \dfrac{\vec{\Lambda}_k}{t}$,
then the column $i$ in $\hat{\vec{A}_{k}} $ is given as

\begin{equation}
[\hat{\vec{A}_{k}} ]_{:,i} = \left \{
    \begin{aligned}
        &\frac{\norm{[\vec{C}]_{:,i}}_2 - \frac{\lambda_1}{t}}{\norm{[\vec{C}]_{:,i}}_2}[\vec{C}]_{:,i}, \ \ &&\textrm{if} \norm{[\vec{C}]_{:,i}}_2 > \frac{\lambda_1}{t}; \\ 
        &0, &&\textrm{otherwise}.
    \end{aligned}
    \right .
\label{eq:A-sol}
\end{equation}

For Problem \eqref{eq:sub-B}, the optimal solution is given by

\begin{equation}
   \hat{\vec{B}_{k}}  = \textrm{prox}_{\frac{\lambda_2}{t}\norm{\cdot}_{*}}(\vec{M}_k -\hat{\vec{A}_{k}} - \frac{\vec{\Lambda}_k}{t}). \label{eq:B-prox}
\end{equation}

The explicit representation of \Cref{eq:B-prox} can be obtained based on SVT operator $\mathcal{\vec{D}}_{\tau}$ \cite{OPT-SIOPT2010-Cai}.
Let $\vec{D}=\vec{M}_k -\hat{\vec{A}_{k}} - \dfrac{\vec{\Lambda}_k}{t}$.
We perform singular value decomposition on $\vec{D}$ such that $\vec{D}=\vec{U}\vec{\Sigma}\vec{V}$, 
where $\vec{\Sigma} = \textrm{diag}(\{\sigma_i\}_{1\leq i\leq r})$ and $\sigma_i$ is the $i$-th largest singular value.
Then $\hat{\vec{B}_{k}} $ is given by

\begin{equation}
\hat{\vec{B}_{k}} = \vec{U}\mathcal{\vec{D}}_{\frac{\lambda_2}{t}}(\vec{\Sigma})\vec{V}
\end{equation}
where $\mathcal{\vec{D}}_{\frac{\lambda_2}{t}}(\vec{\Sigma}) = \textrm{diag}(\{(\sigma_i-\frac{\lambda_2}{t})_+\}).
\label{eq:B-sol}$

For Problem \eqref{eq:sub-M}, it is non-trivial to derive the closed-form of the optimal solution of the sub-problem with respect to $\vec{M}$ since $r(\cdot)$ is a piecewise linear function. 
However, the function is continuous and convex so that we can approach the optimal solution of $\vec{M}$ iteratively by applying gradient-based method.
In our implementation, we apply stochastic gradient descent (SGD) to solve it, and set learning rate as $10^{-3}$ and momentum as 0.9.

Up to now we are extending the classical ADMM to a three-block separable convex programming.
This direct extension, however, is not necessarily convergent, as shown in the previous works \cite{OPT-COA2018-He,OPT-MTPR2016-Chen}. 
To address this issue, a simple correction step was proposed in \cite{OPT-COA2018-He}, shown as follows.

\begin{equation}
    \begin{aligned}
    \begin{pmatrix}
    \vec{B}_{k+1} \\ \vec{M}_{k+1} \\ \vec{\Lambda}_{k+1} 
    \end{pmatrix}
    = 
    \begin{pmatrix}
    \vec{B}_{k} \\ \vec{M}_{k} \\ \vec{\Lambda}_{k}
    \end{pmatrix}
    - \alpha
    \begin{pmatrix}
        \vec{I} & (\tau-1)\vec{I} & \vec{O} \\
        \tau \vec{I} & \vec{I} & \vec{O} \\
        \vec{O} & \vec{O} & \vec{I}
    \end{pmatrix} 
    \begin{pmatrix}
    \vec{B}_{k} - \hat{\vec{B}_{k}} \\ \vec{M}_{k} -\hat{\vec{M}_{k}} \\ \vec{\Lambda}_{k} -\hat{\vec{\Lambda}_{k}} 
    \end{pmatrix},
    \end{aligned}
    \label{eq:correction}
\end{equation}
where $\vec{O}$ denotes zero matrix. $\tau$ is set to $\frac{1}{2}$. $\alpha$ is set to $\frac{3}{4}$.
With this correction step, the extended ADMM can ensure the global convergence. 

The overall optimization procedure is summarized in \Cref{alg:admm}. 
It starts with an initialization for all the variables and hyper-parameters (\cref*{alg:init}). 
Then these variables are updated alternatively in each iteration based on the equations or SGD algorithm, as described above (\cref*{alg:iter-begin} -- \cref*{alg:iter-end}).
Each iteration ends up with a correction step presented as \Cref{eq:correction} (\cref*{alg:correction}). 
The entire optimization procedure exits when the pre-defined condition is satisfied.

\begin{algorithm}[!ht]
    \caption{ADMM for solving Problem \eqref{eq:3-vars}}
    \label{alg:admm}
    
    \begin{algorithmic}[1]
        \Require Feature maps $\vec{Y}_i, \vec{X}_i$, $i=1\cdots N$, given $\lambda_1$, $\lambda_2$.
        \Ensure Structured sparse matrix $\vec{A}$ \& low-rank matrix $\vec{B}$.
        \State Initialize  $k \gets 0$, $\vec{\Lambda}_0$, $\vec{A}_0$, $\vec{B}_0$, $\vec{M}_0$, error tolerance $\epsilon$, $t$; \label{alg:init}
        \While {not converged} 
        \State Calculate $\hat{\vec{A}_{k}}$ by \Cref{eq:A-sol}; \label{alg:iter-begin}
        \State Calculate $\hat{\vec{B}_{k}}$ by \Cref{eq:B-sol};
        \State Calculate $\hat{\vec{M}_{k}}$ by solving Problem \eqref{eq:sub-M} with SGD method;
        \State $\hat{\vec{\Lambda}_{k}} \gets \vec{\Lambda}_k + t(\hat{\vec{A}_{k}} + \hat{\vec{B}_{k}}-\hat{\vec{M}_{k}})$; \label{alg:iter-end}
        \State Perform correction step by \Cref{eq:correction}; \label{alg:correction}
        \State $k\gets k+1$;
        \EndWhile
        \State \Return $\vec{A}_k$ and $\vec{B}_k$;
    \end{algorithmic}   

\end{algorithm}

\subsection{Convergence Analysis}

In this subsection, we prove the convergence of \Cref{alg:admm}. 
Let $f_{1}(\vec{M})=\sum^{N}_{i=1}\norm{\vec{Y}_{i}-r(\vec{MX}_{i})}^{2}_{F}$, $\;f_{2}(\vec{A})=\lambda_{1}\norm{\vec{A}}_{2,1}$, and $\;f_{3}(\vec{B})=\lambda_{2}\norm{\vec{B}}_{*}$. 
Let $\mathbf{m}$ denote the vectorization of $\vec{M}$, i.e., $\mathbf{m}=\textrm{vec}(\vec{M})$, 
and similarly, let $\mathbf{a}=\textrm{vec}(\vec{A})$, and $\mathbf{b}=\textrm{vec}(\vec{B})$.

Using these notations, the problem in \Cref{eq:3-vars} takes the following generic form

\begin{align}\label{eq:3-vars-general}
    \begin{aligned}
        \min_{\vec{A,\,B,\,M}}\; & f_{1}(\vec{M})+f_{2}(\vec{A})+f_{3}(\vec{B}), \\
        \textrm{s.t.}\ \ \      & \vec{C}_{1}\mathbf{a}+\vec{C}_{2}\mathbf{b}-\vec{C}_{3}\mathbf{m}=\mathbf{c},
    \end{aligned}
\end{align}
where $\vec{C}_{1}$, $\vec{C}_{2}$, and $\vec{C}_{3}$ are the identity matrices, and $\vec{c}=\vec{0}$. 
The convergence of ADMM for solving the standard form (\ref{eq:3-vars-general}) was studied in \cite{OPT-COA2018-He,OPT-MTPR2016-Chen}. 
We establish the convergence of our algorithm by transforming the problem in \Cref{eq:3-vars} into a standard form (\ref{eq:3-vars-general}). 
Note that our algorithm alternates between three blocks of variables, $\vec{A}$, $\vec{B}$ and $\vec{M}$.
According to the definitions of $f_{1}(\vec{M})$, $f_{2}(\vec{A})$, and $f_{3}(\vec{B})$, it is easy to verify the problem in \Cref{eq:3-vars} and our algorithm satisfy the convergence conditions of the problem in \Cref{eq:3-vars-general}, as stated in \cite{OPT-COA2018-He}. 
Thus, we have the following theorem.

\begin{mytheorem}
    \label{them1}
    Consider the problem in \Cref{eq:3-vars}, where $f_{1}(\vec{M})$, $f_{2}(\vec{A})$, and $f_{3}(\vec{B})$, are convex functions, and $\vec{C}_{1}$, $\vec{C}_{2}$, and $\vec{C}_{3}$ are the identity matrices, and have full column rank. 
    The sequence $\{\vec{A}_k,\vec{B}_k,\vec{M}_k\}$ generated by \Cref{alg:admm} converges to the optimal solution $\{\vec{A}^{*}, \vec{B}^{*},\vec{M}^{*}\}$ of the problem in \Cref{eq:3-vars}.
\end{mytheorem}

\section{Experimental Results}
\label{sec:results}

\subsection{Experimental Setup}

\Cref{alg:admm} takes in the input and output feature maps generated from the inference on some sample data,
and outputs the approximated network layers. The tested CNNs include \emph{VGG-16} \cite{IMGC-ICLR2015-VGG}, \emph{NIN} \cite{IMGC-ICLR2014-NIN}, \emph{AlexNet} \cite{IMGC-NIPS2012-AlexNet} and \emph{GoogLeNet} \cite{IMGC-CVPR2015-GoogleNet}.
For each network, we first obtain its approximation, and then fine-tune the network based on the obtained structures to restore the accuracy. During the approximation, different layers use different weight coefficients $\lambda_{1}$ and $\lambda_{2}$. 
In our experiments, we find out setting $\lambda_{2}$ to be $2.5 \sim 3$ times larger than $\lambda_{1}$ gives good trade-off between accuracy and model compression rate. And we let
$\lambda_{1}$ ranges from $0.08 \sim 0.3$.
The penalty parameter $t$ in (5) is set to $10^{-3}$.
The runtime of \Cref{alg:admm} varies from layer to layer, ranging from 10 minutes to half an hour.

The inference is conducted on Caffe \cite{DL-ACMMM2014-Caffe} using CIFAR-10 and ILSVRC-2012, i.e., ImageNet \cite{IMGC-CVPR2009-ImageNet}.
After the network approximation, a small initial learning rate of $10^{-5}$ is used in the  fine-tuning step. 
We use three metrics for evaluation, including accuracy loss, compression rate (CR) and speedup ratio. 
The CR is calculated as 

\begin{equation}
	\textrm{CR} = \frac{\textrm{Approximated layer size}}{\textrm{Original layer size}} \times 100\%.
\end{equation}
The accuracy loss is the degradation on accuracy after approximation, denoted by ``\emph{accu. $\bm\downarrow$}'' in the table. 
The ``speed-up'' ratio indicates the acceleration for inference. 


\subsection{Experiments on \emph{CIFAR-10}}

\begin{table}[!tb]
	\centering
	\caption{Results on \emph{VGG-16} with CIFAR-10}
	\label{tab:vgg-layer}
	\resizebox{8.6cm}{!}{
		\begin{tabular}{cccc}
			\toprule
			Layer    & $CR(\bm{A}) (\%)$ & $CR(\bm{B}) (\%)$ & $CR(\bm{A}+\bm{B}) (\%)$ \\ \midrule
			\texttt{conv1-1} &  0.0     & 100.0          & 100     \\ 
			\texttt{conv1-2} &  5.4     & 52.1  & 57.5  \\ 
			\texttt{conv2-1} &  5.4     & 38.2  & 43.6  \\ 
			\texttt{conv2-2} &  2.2     & 28.6  & 30.8  \\ 
			\texttt{conv3-1} &  2.8     & 47.7  & 50.5  \\ 
			\texttt{conv3-2} &  4.0     & 54.3  & 58.3  \\ 
			\texttt{conv3-3} &  10.0    & 59.0  & 69.0  \\ 
			\texttt{conv4-1} &  2.0     & 16.0  & 18.0 \\ 
			\texttt{conv4-2} &  2.0     & 22.4  & 24.4 \\ 
			\texttt{conv4-3} &  4.0     & 16.3  & 24.3 \\ 
			\texttt{conv5-1} &  2.0     & 9.8   & 11.8 \\ 
			\texttt{conv5-2} &  2.0     & 8.7   & 10.7 \\ 
			\texttt{conv5-3} &  2.0     & 6.7   & 8.7  \\ 
			\texttt{fc1}     &  44.2    & 0.0   & 44.2 \\
			\texttt{fc2}     &  36.2    & 0.0   & 36.2 \\
			\texttt{fc3}     &  24.0    & 0.0   & 24.0 \\ \midrule
			CR & \multicolumn{3}{c}{22.5\% ~ (4.44$\times$ reduction of model size)} \\ 
			Speed-up  & \multicolumn{3}{c}{2.2$\times$}     \\ 
			Accu. $\bm\downarrow$ & \multicolumn{3}{c}{0.40\%}           \\
			\bottomrule
		\end{tabular}
	}
\end{table}

\begin{table}[!tb]
	\centering
	\caption{Results on \emph{NIN} with CIFAR-10}
	\label{tab:nin-layer}
	\resizebox{8.6cm}{!}{
		\begin{tabular}{cccc}
			\toprule
			Layer    & $CR(\bm{A}) (\%)$ & $CR(\bm{B}) (\%)$ & $CR(\bm{A}+\bm{B}) (\%)$ \\ \midrule
			\texttt{conv1} &  0.0     & 18.4            & 18.4     \\ 
			\texttt{cccp1} &  0.0     & 100.0          & 100     \\ 
			\texttt{cccp2} &  0.0     & 100.0          & 100     \\ 
			\texttt{conv2} &  0.0     & 16.9           & 16.9  \\ 
			\texttt{cccp3} &  0.0     & 100.0          & 100     \\ 
			\texttt{cccp4} &  0.0     & 100.0          & 100     \\ 
			\texttt{conv3} &  0.0     & 38.2           & 38.2 \\ 
			\texttt{cccp5} &  0.0     & 100.0          & 100     \\ 
			\texttt{cccp6} &  0.0     & 100.0          & 100     \\  \midrule
			CR & \multicolumn{3}{c}{36.0\% ~ (2.77$\times$ reduction of model size)} \\ 
			Speed-up  & \multicolumn{3}{c}{2.2$\times$}     \\ 
			Accu. $\bm\downarrow$ & \multicolumn{3}{c}{0.41\%}           \\
			\bottomrule
		\end{tabular}
	}
\end{table}

\subsubsection{VGG-16}
\emph{VGG-16} \cite{IMGC-ICLR2015-VGG} network is a convolutional neural network consisting of 13 convolution layers and 3 FC layers.
All the convolutional filters have the same spatial size of $3 \times 3$. 
We test the proposed method with experiments on the CIFAR-10 dataset which consists of 50K training images and 10K test images. 
We first train a \emph{VGG-16} network from scratch to obtain the baseline, which has an accuracy of 92.05\%.
To make the approximation, 1000 images are selected from training set for inference and the input and output feature maps are collected for \Cref{alg:admm}.


The approximation is performed on each layer sequentially. 
The layer-wise approximation results are shown in \Cref{tab:vgg-layer}. 
In our experiment, we find out approximating the first convolutional layer may lead to significant accuracy drop.
Therefore, the first layer is not approximated.
The sparse component $\vec{A}$ is stored in CSR format. 
Moreover, we constrain the sparse component $\vec{A}$ to be structured sparse to accelerate the computation as in  \cite{SPEED-NIPS2016-Wen}. 
The low-rank component is represented by the product of two smaller matrices.
For FC layers, we only use the sparse component for approximation to reduce accuracy drop.

The performance comparison with other previous work \cite{SPEED-ICLR2017-Li} is presented in \Cref{tab:comparison-cifar}.
With the approximation, the model size is reduced by $4.44\times$, which corresponds to  $2.2\times$ speedup on inference. 
Both compression rate and speedup ratio outperform \cite{SPEED-ICLR2017-Li}. 
Without fine-tuning, there is some classification accuracy drop.
In order to restore the accuracy of the compressed model, we retrain the compressed network with the training set for 5 epochs. 
With this fine-tuning step the accuracy loss reduces from 1.8\% to only 0.40\%, which becomes very close to the accuracy of the original VGG-16.

\begin{table}[tb!]
	\centering
	\caption{Comparison on CIFAR-10}
	\renewcommand{\arraystretch}{1.5}
	\begin{tabular}{c|c|c|c|c}
		\toprule
		Model                    & Method          & Accu. $\bm\downarrow$  & CR            & Speed-up      \\ \midrule
		\multirow{3}{*}{VGG-16}  & Original        & 0.00\%                    & 1.00          & 1.00          \\ 
                                 & ICLR'17 \cite{SPEED-ICLR2017-Li}  & \textbf{0.06\%}           & 2.70          & 1.80          \\ 
		                         & Ours            & 0.40\%                    & \textbf{4.44} & \textbf{2.20} \\ \midrule        
		\multirow{3}{*}{NIN}     & Original        & 0.00\%                    & 1.00          & 1.00          \\ 
                                 & ICLR'16 \cite{SPEED-ICLR2016-Tai}         & 1.43\%                    & 1.54          & 1.50          \\ 
                                 & IJCAI'18 \cite{SPEED-IJCAI2018-Kas}       & 1.43\%                    & 1.45          & -             \\ 
		                         & Ours            & \textbf{0.41\%}           & \textbf{2.77} & \textbf{1.70} \\ \midrule                   
	\end{tabular}
	\label{tab:comparison-cifar}
\end{table}

\begin{figure}[!tb]
	\centering
	\pgfplotsset{
	width =0.72\linewidth,
	height=0.52\linewidth
}

\begin{tikzpicture}
\begin{axis}[
    ybar,
    xlabel={Number of layers approximated},
    xtick={2,...,16},
    ylabel={Accu.~Drop (\%)},
    ylabel near ticks,
    bar width = 5pt,
    ymin=0,
    ymax=4.2,
    legend style={at={(0.5,1.25)},
    draw=none,anchor=north,legend columns=2},
]
\addplot +[ybar, fill=myblue, draw=black, area legend] table [x={alpha},  y={accu-asym}] {accu-cifar.dat};
\addplot +[fill=mygreen, draw=black, area legend]  table [x={alpha},  y={accu-sym}]   {accu-cifar.dat};
\legend{Asymmetric, Symmetric}
\end{axis}
\end{tikzpicture}
	\caption{Accuracy on CIFAR-10 using symmetric reconstruction and asymmetric reconstruction.}
	\label{fig:asym-accuracy}
\end{figure}
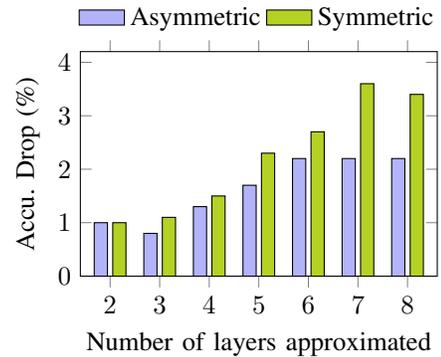

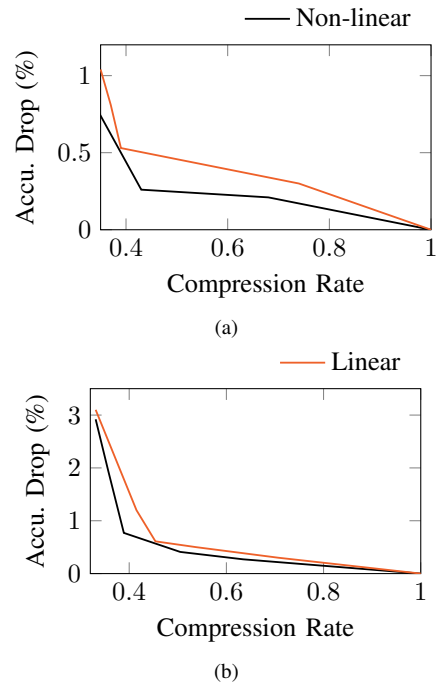
\begin{figure}[!tb]
    \centering
    \subfloat[]{\pgfplotsset{
    width =0.68\linewidth,
    height=0.46\linewidth
}
\begin{tikzpicture}[scale=1]
\begin{axis}[minor tick num=0,
    xmin=0.35, xmax=1,
    ymin=0, ymax= 1.2,
    xlabel={Compression Rate},
    ylabel={Accu.~Drop (\%)},
    y label style={at={(-3.4,0.5)}},
    xlabel near ticks,
    ylabel near ticks,
    legend style={
        draw=none,
    at={(0.98,1.26)},
    anchor=north east,
    legend columns=1,
}
]
\addplot +[line width=0.7pt] [color=black,    no marks] table [x={relu_x},   y={relu_y}]   {nonlinear.dat};
\addplot +[line width=0.7pt] [color=myorange, no marks] table [x={norelu_x}, y={norelu_y}] {nonlinear.dat};
\legend{{Non-linear}}
\end{axis}
\end{tikzpicture}} 
    
    \subfloat[]{\pgfplotsset{
    width =0.68\linewidth,
    height=0.46\linewidth
}
\begin{tikzpicture}[scale=1]
\begin{axis}[minor tick num=0,
    xmin=0.32, xmax=1,
    ymin=0, ymax= 3.5,
    yticklabel style={/pgf/number format/.cd, fixed, fixed zerofill, precision=0, /tikz/.cd},
    xlabel={Compression Rate},
    ylabel={Accu.~Drop (\%)},
    y label style={at={(-0.2,0.5)}},
    xlabel near ticks,
    ylabel near ticks,
    legend style={
        draw=none,
    at={(0.98,1.26)},
    anchor=north east,
    legend columns=1,
}
]
\addplot +[line width=0.7pt] [color=black,    no marks] table [x={relu_x},   y={relu_y}]   {nonlinear.dat};
\addplot +[line width=0.7pt] [color=myorange, no marks] table [x={norelu_x}, y={norelu_y}] {nonlinear.dat};
\legend{,{Linear}}
\end{axis}
\end{tikzpicture}}
    \caption{Comparison of reconstructing linear response and non-linear response:
    (a) layer \texttt{conv2-1}; (b) layer \texttt{conv3-1}.}
    \label{fig:relu}
\end{figure}

If a shallow layer is approximated, the approximation error may be accumulated when deeper layers are approximated.
In order to handle this issue, we take the `asymmetric' strategy used in \cite{SPEED-CVPR2015-Zhang}. 
We approximate the layers from shallow to deep.
When approximating a deep layer, use the response produced by all previous layers instead of the non-approximate response as the input feature map $\bm{X}_i$.
\Cref{fig:asym-accuracy} shows the comparison of classification error increase.
We can observe that with more layers being approximated, the performance becomes worse for both strategies.
However, the asymmetric version loses less accuracy.

\begin{figure}[!tb]
    \centering
    \begin{minipage}{.34\linewidth}
        \centering
        \subfloat[]{\includegraphics[width=.98\linewidth]{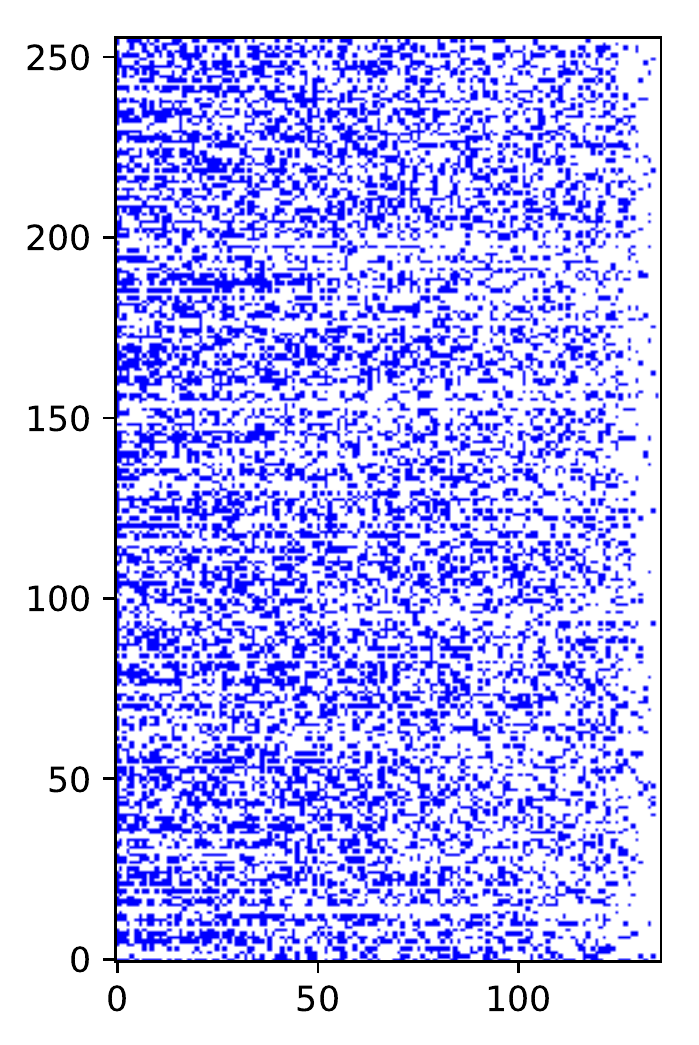}}
    \end{minipage}
    \begin{minipage}{.64\linewidth}
        \centering
        \subfloat[]{\includegraphics[width=.98\linewidth]{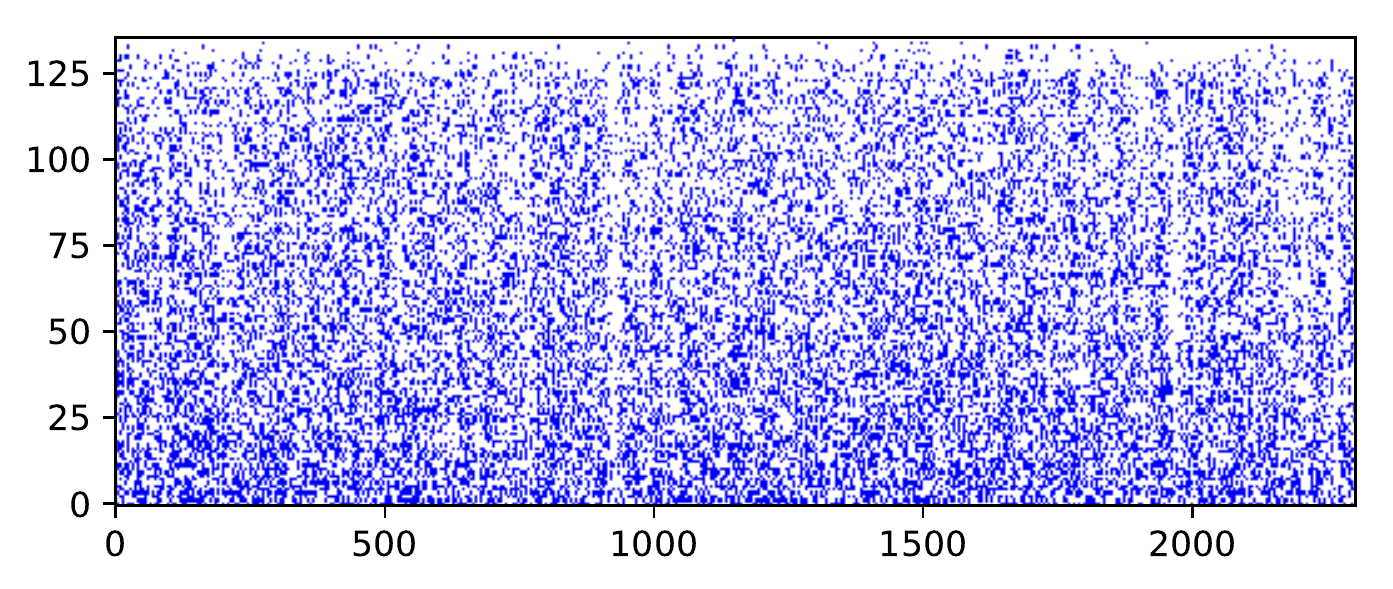}}\\
        \subfloat[]{\includegraphics[width=.88\linewidth]{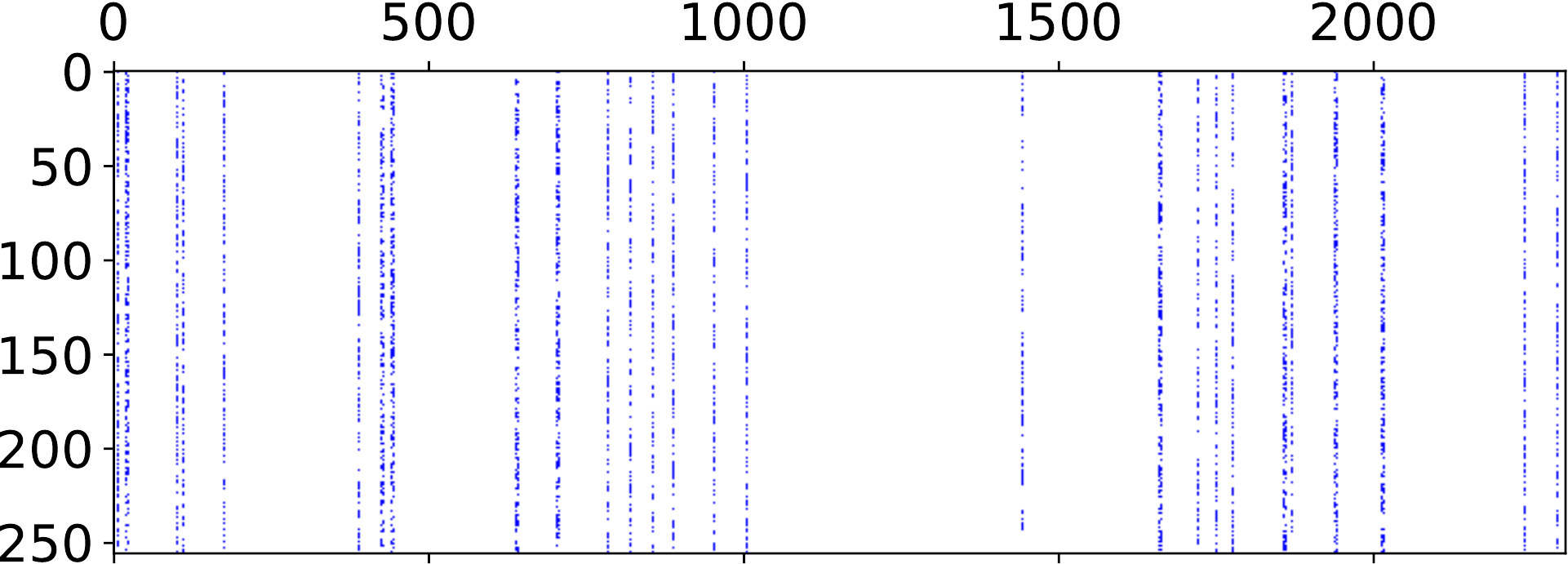}}
    \end{minipage}
    \caption{Approximated filters of \texttt{conv3-1}. Blue dots have non-zero values.
        Low-rank filter $\vec{B}$ with rank $136$ is decomposed into $\vec{UV}$, both of which have rank $136$.
        (a) Matrix $\vec{U}$;
        (b) Matrix $\vec{V}$.
        (c) Column-wise sparse filter $\vec{A}$.
    } 
    \label{fig:visualize}
\end{figure}

We further compare the performance between reconstructing non-linear response and reconstructing linear response. 
We perform the comparison on a single layer each time, while the remaining layers are kept unchanged. 
In \Cref{fig:relu}, we plot the relation between the CR and the accuracy degradation of two approaches of different layers.
The performance is evaluated by the accuracy drop compared with original model.
We take two convolutional layers in two different stages of the \emph{VGG-16}, including \texttt{conv2-1} and \texttt{conv3-1}.
\Cref{fig:relu} shows that under the same CR, reconstructing non-linear response achieves lower accuracy drop than reconstructing linear response,
which verifies the advantage of reconstructing the non-linear response.
In \Cref{fig:visualize}, we visualize the sparse filter and low-rank filter after the approximation of layer \texttt{conv3-1}.
$\vec{B}$ has rank 136 and it can be further decomposed by $\vec{B}=\vec{UV}$,
where both $\vec{U}$ and $\vec{V}$ have rank 136.

\subsubsection{NIN}
Network-in-network (\emph{NIN}) \cite{IMGC-ICLR2014-NIN} has 9 convolutional layers among which 6 layers have a spatial size of $1\times 1$.
Considering that these $1\times 1$ convolutional layers have less contribution to the overall model size and computation, we focus on remaining three layers which have spatial size of $3\times 3$ or $5 \times 5$. 
We present the layer-wise approximation results in \Cref{tab:nin-layer}. 
It can be observed that for all the approximated convolutional layers, only low-rank component is used and structured sparse didn't show up, which means approximating \emph{NIN} using CIFAR10 dataset reduces to low-rank approximation and sparse components are not beneficial to the objectives.  
It indicates that the proposed unified framework is flexible to find good solutions and does not rely on prior assumptions to achieve good results. 

Experimental results using the same network (i.e., \emph{NIN}) and CIFAR10 are reported in previous work \cite{SPEED-ICLR2016-Tai,SPEED-IJCAI2018-Kas}. 
The comparison of accuracy loss, compression rates and the accuracy is shown in \Cref{tab:comparison-cifar}. 
We can see that the number of parameters is reduced by  2.77$\times$ and the inference time is accelerated by 1.70$\times$, with only 0.41\% accuracy loss compared with original model. 
All these three metrics are significantly better than previous work \cite{SPEED-ICLR2016-Tai,SPEED-IJCAI2018-Kas}.

\subsection{Experiments on \emph{ImageNet}}
\subsubsection{AlexNet}
\emph{AlexNet} \cite{IMGC-NIPS2012-AlexNet} has 5 convolutional layers and 3 FC layers. 
It is tested for the ImageNet classification task. We evaluate the top-5 accuracy with single-view.
The ILSVRC-2012 dataset consists of 1.2 million training images and 50 thousand test images. 
Images are resized with 256 pixels on the shorter side.
The testing image is on the center crop of 224 $\times$ 224 pixels.
We use the pre-trained model provided by Caffe Model Zoo as the baseline. 
In our experiment, we first select 1500 images from the training set and collect their responses for building the approximate network. 
The layer-wise approximation results are demonstrated in \Cref{tab:alexnet-layer}. 
The first two convolutional layers of \emph{AlexNet} are not approximated, in order to preserve good accuracy. 
For FC layers, again we only use the structured sparse component for approximation. 

The compression rates and the accuracy comparison are shown in \Cref{tab:comparison-imagenet}. 
From the table we see that the network is compressed by more than 5$\times$, which outperforms \cite{SPEED-ICLR2016-Tai},
\cite{SPEED-ICLR2016-Kim}, and \cite{SPEED-CVPR2018-Yu}, while the top-5 accuracy drop is only 1.3\%.
This reveals that the proposed approximation framework can remarkably compress \emph{AlexNet} while keeping good accuracy. 

\begin{table}[!tb]
	\caption{Results on \emph{AlexNet} with ILSVRC-2012}
	\centering
	\resizebox{8.6cm}{!}{
		\begin{tabular}{cccc}
			\toprule
			Layer    & $CR(\bm{A}) (\%)$ & $CR(\bm{B}) (\%)$ & $CR(\bm{A}+\bm{B}) (\%)$ \\ \midrule
			\texttt{conv1} & 0.0          & 100.0     & 100.0     \\ 
			\texttt{conv2} & 21.4          & 23.6    & 45.0     \\ 
			\texttt{conv3} & 30.0   &  22.9   & 52.9   \\ 
			\texttt{conv4} & 0.0    &  32.6   & 32.6  \\ 
			\texttt{conv5} & 0.0    &  26.0   & 26.0  \\ 
			\texttt{fc1}       & 12.8 & 0.0  & 12.8  \\ 
			\texttt{fc2}       & 26.2 & 0.0  & 26.2   \\
			\texttt{fc3}       & 18.8 & 0.0  & 18.8   \\ \midrule
			CR  & \multicolumn{3}{c}{18.0\% ~ (5.56$\times$ reduction of model size)} \\ 
			 Speed-up   & \multicolumn{3}{c}{1.1$\times$}    \\ 
			Top-5 accu. $\bm\downarrow$ & \multicolumn{3}{c}{1.27\%}           \\
			\bottomrule
		\end{tabular}
	}
	\label{tab:alexnet-layer}
\end{table}

\subsubsection{GoogLeNet}
\emph{GoogLeNet} \cite{IMGC-CVPR2015-GoogleNet} is another widely used network in image recognition and classification. 
Different from \emph{AlexNet}, \emph{GoogLeNet} combines two spatial sizes of convolutional filters, $3 \times 3$ and $5 \times 5$, in each inception block.
In order to collect the input samples for optimization, we use a pre-trained model provided by Caffe Model Zoo to perform inference and dump the input and output feature maps of each convolutional layer.
After performing approximation on \emph{GoogLeNet}, both model size and inference time are reduced.
The layer-wise approximation results are shown in \Cref{tab:googlenet-layer}. 
The comparison of accuracy loss, compression rates and accuracy are shown in \Cref{tab:comparison-imagenet}. 
We can see that the model size is reduced by  2.87$\times$ and the inference time is accelerated by 1.35$\times$, without loss on accuracy.
All these three metrics are significantly better than previous works using the same network model and dataset  \cite{SPEED-ICLR2016-Tai,SPEED-ICLR2016-Kim,SPEED-CVPR2018-Yu}.

\begin{table}[!tb]
	\centering
	\caption{Results on \emph{GoogLeNet} with ILSVRC-2012}
	\resizebox{8.6cm}{!}{
		
		\begin{tabular}{cccc}
			\toprule
			Layer    & $CR(\bm{A}) (\%)$ & $CR(\bm{B}) (\%)$ & $CR(\bm{A}+\bm{B}) (\%)$ \\ \midrule
			\texttt{conv1}        & 0.0        & 100.0     & 100.0  \\ 
			\texttt{conv2}        & 0.0        & 100.0     & 100.0  \\ 
			\texttt{inception-3a} & 2.1        & 31.1      & 33.2 \\ 
			\texttt{inception-3b} & 5.4        & 39.8      & 45.3 \\ 
			\texttt{inception-4a} & 3.8        & 28.6      & 32.4 \\ 
			\texttt{inception-4b} & 2.3        & 23.7      & 26.1 \\ 
			\texttt{inception-4c} & 8.9        & 29.9      & 38.9 \\ 
			\texttt{inception-4e} & 2.7        & 23.7      & 26.5 \\ 
			\texttt{inception-5a} & 2.4        & 28.8      & 31.2 \\ 
			\texttt{inception-5b} & 1.6        & 31.6      & 33.3 \\ 
			\texttt{fc}           & 35.0       & 0.0       & 35.0 \\ \midrule
			CR & \multicolumn{3}{c}{34.8\% ~ ($2.87\times$ reduction of model size)} \\ 
			Speed-up  & \multicolumn{3}{c}{$1.35\times$}     \\ 
			Top-5 accu. $\bm\downarrow$ & \multicolumn{3}{c}{0.00\%}           \\
			\bottomrule
		\end{tabular}
	}
	\label{tab:googlenet-layer}
	
\end{table}

\begin{table}[!tb]
	\centering
	\caption{Comparison on ILSVRC-2012}
	\resizebox{8.6cm}{!}{
	\renewcommand{\arraystretch}{1.25}
	\small
	\begin{tabular}{c|c|c|c|c}
		\toprule
		Model                    &  Method   & Top-5 Accu.$\bm\downarrow$ & CR & Speed-up  \\ \midrule
		\multirow{4}{*}{AlexNet} &  Original & 0.00\%           & 1.00          & 1.00  \\ 
        &  ICLR'16 \cite{SPEED-ICLR2016-Tai} & \textbf{0.37\%}  & 5.00          & \textbf{1.82}  \\ 
        &  ICLR'16 \cite{SPEED-ICLR2016-Kim} & 1.70\%           & 5.46          & 1.81  \\
        &  CVPR'18 \cite{SPEED-CVPR2018-Yu} & 1.43\%           & 1.50          & -     \\
		&  Ours     & 1.27\%           & \textbf{5.56} & 1.10  \\ \midrule
		\multirow{4}{*}{GoogleNet}  &  Original & 0.00\%           & 1.00          & 1.00  \\ 
		&  ICLR'16 \cite{SPEED-ICLR2016-Tai} & 0.42\%           & 2.84          & 1.20  \\ 
		&  ICLR'16 \cite{SPEED-ICLR2016-Kim} & 0.24\%           & 1.28          & 1.23  \\
		&  CVPR'18 \cite{SPEED-CVPR2018-Yu}  & 0.21\%           & 1.50          & -     \\
		&  Ours     & \textbf{0.00\%}  & \textbf{2.87} & \textbf{1.35}  \\ \midrule                        
	\end{tabular}                                                             
	}
	\label{tab:comparison-imagenet}
\end{table}



\section{Conclusion}
\label{sec:conclu}

In this paper, we have proposed a unified approximation model for deep neural networks with simultaneous low-rank approximation and structured sparsification.
It also considers the non-linear activation to retain the accuracy. 
To obtain this model, a layer-wise optimization problem is presented, relaxed, and solved with an extended ADMM algorithm whose convergence is provably guaranteed. 
The effectiveness of the proposed approximation framework is verified on \emph{VGG-16}, \emph{NIN}, \emph{GoogLeNet} and \emph{AlexNet}.
By sacrificing little accuracy, \emph{VGG-16} and \emph{AlexNet} are compressed by up to 5.56$\times$.
\emph{GoogLeNet} is compressed by nearly 3$\times$ without loss of accuracy. 
What's more, since structured sparse filters and low-rank filters are independent to each other, more inference speedup may be expected if taking actual architecture and parallel computing into account.

\section{Acknowledgment}

This work is partially supported by
Tencent Technology,
The Research Grants Council of Hong Kong SAR (Project No.~CUHK24209017),
Beijing National Research Center for Information Science and Technology (BNR2019ZS01001), 
and NSFC under grant (No.~61872206, No.~61976164).

{
    \balance
    \bibliographystyle{IEEETran}
    \bibliography{./ref/Top-sim,./ref/LEARN,./ref/DL,./ref/SPEED,./ref/CV,./ref/addition}

\begin{thebibliography}{10}
\providecommand{\url}[1]{#1}
\csname url@samestyle\endcsname
\providecommand{\newblock}{\relax}
\providecommand{\bibinfo}[2]{#2}
\providecommand{\BIBentrySTDinterwordspacing}{\spaceskip=0pt\relax}
\providecommand{\BIBentryALTinterwordstretchfactor}{4}
\providecommand{\BIBentryALTinterwordspacing}{\spaceskip=\fontdimen2\font plus
\BIBentryALTinterwordstretchfactor\fontdimen3\font minus
  \fontdimen4\font\relax}
\providecommand{\BIBforeignlanguage}[2]{{%
\expandafter\ifx\csname l@#1\endcsname\relax
\typeout{** WARNING: IEEEtran.bst: No hyphenation pattern has been}%
\typeout{** loaded for the language `#1'. Using the pattern for}%
\typeout{** the default language instead.}%
\else
\language=\csname l@#1\endcsname
\fi
#2}}
\providecommand{\BIBdecl}{\relax}
\BIBdecl

\bibitem{SPEED-NIPS2013-Denil}
M.~Denil, B.~Shakibi, L.~Dinh, N.~De~Freitas \emph{et~al.}, ``Predicting
  parameters in deep learning,'' in \emph{Proc.~NIPS}, 2013, pp. 2148--2156.

\bibitem{SPEED-ICLR2016-Han}
S.~Han, H.~Mao, and W.~J. Dally, ``{Deep Compression}: Compressing deep neural
  networks with pruning, trained quantization and huffman coding,'' in
  \emph{Proc.~ICLR}, 2016.

\bibitem{SPEED-NIPS2016-Wen}
W.~Wen, C.~Wu, Y.~Wang, Y.~Chen, and H.~Li, ``Learning structured sparsity in
  deep neural networks,'' in \emph{Proc.~NIPS}, 2016, pp. 2074--2082.

\bibitem{SPEED-IJCNN2018-Dai}
R.~Dai, L.~Li, and W.~Yu, ``Fast training and model compression of gated {RNN}s
  via singular value decomposition,'' in \emph{Proc.~IJCNN}, 2018.

\bibitem{SPEED-CVPR2015-Zhang}
X.~Zhang, J.~Zou, X.~Ming, K.~He, and J.~Sun, ``Efficient and accurate
  approximations of nonlinear convolutional networks,'' in \emph{Proc.~CVPR},
  2015, pp. 1984--1992.

\bibitem{SPEED-NIPS2015-Novikov}
A.~Novikov, D.~Podoprikhin, A.~Osokin, and D.~Vetrov, ``Tensorizing neural
  networks,'' in \emph{Proc.~NIPS}, 2015, pp. 442--450.

\bibitem{SPEED-ICLR2016-Tai}
C.~Tai, T.~Xiao, Y.~Zhang, X.~Wang \emph{et~al.}, ``Convolutional neural
  networks with low-rank regularization,'' in \emph{Proc.~ICLR}, 2016.

\bibitem{SPEED-NIPS2017-Alvarez}
J.~M. Alvarez and M.~Salzmann, ``Compression-aware training of deep networks,''
  in \emph{Proc.~NIPS}, 2017, pp. 856--867.

\bibitem{SPEED-CVPR2017-Yu}
X.~Yu, T.~Liu, X.~Wang, and D.~Tao, ``On compressing deep models by low rank
  and sparse decomposition,'' in \emph{Proc.~CVPR}, 2017, pp. 7370--7379.

\bibitem{SPEED-ICCV2017-He}
Y.~He, X.~Zhang, and J.~Sun, ``Channel pruning for accelerating very deep
  neural networks,'' in \emph{Proc.~ICCV}, 2017.

\bibitem{DL-ICML2010-Nair}
V.~Nair and G.~E. Hinton, ``Rectified linear units improve restricted boltzmann
  machines,'' in \emph{Proc.~ICML}, 2010, pp. 807--814.

\bibitem{OPT-FTML2011-Boyd}
S.~Boyd, N.~Parikh, E.~Chu, B.~Peleato, J.~Eckstein \emph{et~al.},
  ``Distributed optimization and statistical learning via the alternating
  direction method of multipliers,'' \emph{Foundations and Trends in Machine
  learning}, vol.~3, no.~1, pp. 1--122, 2011.

\bibitem{IMGC-ICLR2015-VGG}
K.~Simonyan and A.~Zisserman, ``Very deep convolutional networks for
  large-scale image recognition,'' in \emph{Proc.~ICLR}, 2015, pp. 1--14.

\bibitem{IMGC-ICLR2014-NIN}
M.~Lin, Q.~Chen, and S.~Yan, ``Network in network,'' \emph{arXiv preprint
  arXiv:1312.4400}, 2013.

\bibitem{IMGC-NIPS2012-AlexNet}
A.~Krizhevsky, I.~Sutskever, and G.~E. Hinton, ``{ImageNet} classification with
  deep convolutional neural networks,'' in \emph{Proc.~NIPS}, 2012, pp.
  1097--1105.

\bibitem{IMGC-CVPR2015-GoogleNet}
C.~Szegedy, W.~Liu, Y.~Jia, P.~Sermanet, S.~Reed, D.~Anguelov, D.~Erhan,
  V.~Vanhoucke, and A.~Rabinovich, ``Going deeper with convolutions,'' in
  \emph{Proc.~CVPR}, 2015, pp. 1--9.

\bibitem{SPEED-ICLR2017-Li}
H.~Li, A.~Kadav, I.~Durdanovic, H.~Samet, and H.~P. Graf, ``Pruning filters for
  efficient convnets,'' in \emph{Proc.~ICLR}, 2017.

\bibitem{SPEED-CVPR2018-Yu}
R.~Yu, A.~Li, C.-F. Chen, J.-H. Lai, V.~I. Morariu, X.~Han, M.~Gao, C.-Y. Lin,
  and L.~S. Davis, ``{NISP}: Pruning networks using neuron importance score
  propagation,'' in \emph{Proc.~CVPR}, 2018.

\bibitem{SPEED-BMVC2014-Jaderberg}
M.~Jaderberg, A.~Vedaldi, and A.~Zisserman, ``Speeding up convolutional neural
  networks with low rank expansions,'' in \emph{Proc.~BMVC}, 2014.

\bibitem{SPEED-ICLR2015-Lebedev}
V.~Lebedev, Y.~Ganin, M.~Rakhuba, I.~Oseledets, and V.~Lempitsky, ``Speeding-up
  convolutional neural networks using fine-tuned cp-decomposition,'' in
  \emph{Proc.~ICLR}, 2015.

\bibitem{SPEED-IJCAI2018-Kas}
S.~P. Kasiviswanathan, N.~Narodytska, and H.~Jin, ``Network approximation using
  tensor sketching,'' in \emph{Proc.~IJCAI}, 2018, pp. 2319--2325.

\bibitem{SPEED-ICCV2017-Wen}
W.~Wen, C.~Xu, C.~Wu, Y.~Wang, Y.~Chen, and H.~Li, ``Coordinating filters for
  faster deep neural networks,'' in \emph{Proc.~ICCV}, 2017, pp. 658--666.

\bibitem{SPEED-NIPS2014-Denton}
E.~L. Denton, W.~Zaremba, J.~Bruna, Y.~LeCun, and R.~Fergus, ``Exploiting
  linear structure within convolutional networks for efficient evaluation,'' in
  \emph{Proc.~NIPS}, 2014, pp. 1269--1277.

\bibitem{SPEED-ECCV2016-Zhou}
H.~Zhou, J.~M. Alvarez, and F.~Porikli, ``Less is more: Towards compact cnns,''
  in \emph{Proc.~ECCV}, 2016, pp. 662--677.

\bibitem{ML-TPAMI2013-Liu}
G.~Liu, Z.~Lin, S.~Yan, J.~Sun, Y.~Yu, and Y.~Ma, ``Robust recovery of subspace
  structures by low-rank representation,'' \emph{IEEE TPAMI}, vol.~35, no.~1,
  pp. 171--184, 2013.

\bibitem{OPT-SIOPT2010-Cai}
J.-F. Cai, E.~J. Cand{\`e}s, and Z.~Shen, ``A singular value thresholding
  algorithm for matrix completion,'' \emph{SIAM Journal on Optimization
  (SIOPT)}, vol.~20, no.~4, pp. 1956--1982, 2010.

\bibitem{OPT-COA2018-He}
B.-S. He and X.~Yuan, ``A class of {ADMM}-based algorithms for three-block
  separable convex programming,'' \emph{Computational Optimization and
  Applications}, 2018.

\bibitem{OPT-MTPR2016-Chen}
C.~Chen, B.~He, Y.~Ye, and X.~Yuan, ``The direct extension of {ADMM} for
  multi-block convex minimization problems is not necessarily convergent,''
  \emph{Mathematical Programming}, vol. 155, no. 1-2, pp. 57--79, 2016.

\bibitem{DL-ACMMM2014-Caffe}
Y.~Jia, E.~Shelhamer, J.~Donahue, S.~Karayev, J.~Long, R.~Girshick,
  S.~Guadarrama, and T.~Darrell, ``Caffe: Convolutional architecture for fast
  feature embedding,'' in \emph{Proc.~Multimedia}, 2014, pp. 675--678.

\bibitem{IMGC-CVPR2009-ImageNet}
J.~Deng, W.~Dong, R.~Socher, L.-J. Li, K.~Li, and L.~Fei-Fei, ``{ImageNet}: A
  large-scale hierarchical image database,'' in \emph{Proc.~CVPR}, 2009, pp.
  248--255.

\bibitem{SPEED-ICLR2016-Kim}
Y.-D. Kim, E.~Park, S.~Yoo, T.~Choi, L.~Yang, and D.~Shin, ``Compression of
  deep convolutional neural networks for fast and low power mobile
  applications,'' in \emph{Proc.~ICLR}, 2016.

\end{thebibliography}
}

\end{document}